\documentclass{article}
\usepackage{spconf,amsmath,graphicx}
\usepackage[dvipsnames]{xcolor}

\usepackage{mystyle}
\usepackage[hyphens]{url}

\usepackage{multirow}
\usepackage{subcaption}
\usepackage{float}

\usepackage{tikz}
\usetikzlibrary{calc}
\usetikzlibrary{bayesnet}

\newcommand{\embosi}{Mboshi}
\newcommand{\ligaikuma}{Lig\_Aikuma}

\DeclareMathOperator*{\argmax}{arg\,max}

\title{Bayesian Models for Unit Discovery on a Very Low Resource Language}

\name{Lucas Ondel$^{1}$,  Pierre Godard$^{2}$, Laurent Besacier$^{3}$, Elin Larsen$^{6}$, Mark Hasegawa-
Johnson$^{4}$}
\secondlinename{ Odette Scharenborg$^{5}$\thanks{O. Scharenborg was supported by a Vidi-grant from NWO (grant number: 276-89-003)}, Emmanuel Dupoux$^{6}$, Lukas Burget$^{1}$, Fran\c{c}ois 
Yvon$^{2}$, Sanjeev Khudanpur$^7$}

\address{
1. Brno University of Technology, Brno, Czech Republic,
2. LIMSI, CNRS, Universit\'e Paris Saclay\\
3. LIG, CNRS, Universit\'e Grenoble Alpes,
4. University of Illinois, Urbana, IL, USA\\
5. Centre for Language Studies, Radboud University, Nijmegen, Netherlands\\
6. CoML, ENS/EHESS/PSL Research University/CNRS/INRIA, Paris, France \\
7. Johns Hopkins University, Baltimore, MD USA
}

\begin{document}

\maketitle

\begin{abstract}
Developing speech technologies for low-resource languages has become a very active research field over the
last decade. Among others, Bayesian models have shown some promising results on artificial examples but 
still lack of \emph{in situ} experiments. Our work applies state-of-the-art Bayesian models to unsupervised 
Acoustic Unit Discovery (AUD) in a real low-resource language scenario. We also show that Bayesian models 
can naturally integrate information from other resourceful languages by means of \emph{informative prior} 
leading to more consistent discovered units. Finally, discovered acoustic units are used, either as the 
1-best sequence or as a lattice, to perform word segmentation. Word segmentation results show that this 
Bayesian approach clearly outperforms a Segmental-DTW baseline on the same corpus.

\end{abstract}

\begin{keywords}
Acoustic Unit Discovery, Low-Resource ASR, Bayesian Model, Informative Prior.
\end{keywords}

\section{Introduction}
\label{sec:intro}

Out of nearly 7000 languages spoken worldwide, current speech (ASR, TTS, voice 
search, etc.) technologies barely address 200 of them. Broadening ASR 
technologies to ideally all possible languages is a challenge with very high stakes 
in many areas and is at the heart of several fundamental research problems ranging 
from psycholinguistic (how humans learn to recognize speech) to pure machine learning 
(how to extract knowledge from unlabeled data). The present work focuses on the 
narrow but important problem of unsupervised Acoustic Unit Discovery (AUD). It takes 
place as the continuation of an ongoing effort to develop a Bayesian model suitable 
for this task, which stems from the seminal work of \cite{Lee2012} later refined and 
made scalable in \cite{Ondel2016}. This model, while rather crude, has shown that it can provide a clustering 
accurate enough to be used in topic identification of spoken document in unknown languages \cite{
Kesiraju2017}. It was also shown that this model can be further improved by incorporating a Bayesian 
"phonotactic" language model learned jointly with the acoustic units \cite{Ondel2017}.  Finally, 
following the work in \cite{Johnson2016} it has been combined successfully with variational 
auto-encoders leading to a model combining the potential of both deep neural networks and Bayesian 
models \cite{Ebbers2017}. The contribution of this 
work is threefold:
\begin{itemize}
	\item we compare two Bayesian models (\cite{Ondel2016} and \cite{Ebbers2017}) for acoustic unit discovery (AUD) on a very low resource 
    language speech corpus,
    \item we investigate the use of "informative prior" to improve the performance of 
    Bayesian models by using information from resourceful languages,
    \item as an extrinsic evaluation of AUD quality, we cascade AUD with sequence/lattice based word discovery
    \cite{HeymannW2013}.
\end{itemize}

\section{Models}
\label{sec:models}

The AUD model described in \cite{Lee2012, Ondel2016} is a non-parametric Bayesian Hidden Markov 
Model (HMM). This model is topologically equivalent to a phone-loop model with two major differences:

\begin{itemize}
	\item since it is trained in an unsupervised fashion the elements of the loop cannot directly be
    interpreted as the actual phones of the target language but rather as some acoustic units (defined as 3-states left-to-right sub-HMM) whose
    time scale approximately corresponds the phonetic time
    scale.
    \item to cope with the unknown number of acoustic units needed to properly describe  speech, 
    the model assumes a theoretically infinite number of potential acoustic units. However, during  
    inference, the prior over the weight of the acoustic units (a Dirichlet Process \cite{TehJor2010})
    will act as a sparsity regularizer leading to a model which explains the data with a relatively 
    small number of units.    
\end{itemize}
%
%
%
In this work, we have used two variants of this original model. The first one (called HMM model in the remainder of this paper), following the analysis led in \cite{Kurihara2007},
approximates the Dirichlet Process prior by a mere symmetric Dirichlet 
prior. This approximation, while retaining the sparsity constraint, avoids
the complication of dealing with the variational treatment of the 
stick breaking process frequent in Bayesian non-parametric models. The second 
variant, which we shall denote Structured Variational AutoEncoder (SVAE) AUD, 
is based upon the work of \cite{Johnson2016} and embeds the HMM model into 
the Variational AutoEncoder framework \cite{KingmaW2013}.  A 
very similar version of the SVAE for AUD was developed independently and presented in
\cite{Ebbers2017}. The main noteworthy difference between \cite{Ebbers2017} 
and our model is that we consider a fully Bayesian version of the HMM embedded
in the VAE; and the posterior distribution and the VAE parameters are trained 
jointly using the Stochastic Variational Bayes \cite{Johnson2016,Hoffman2013}. 
%
%
For both variants, the prior over the HMM parameters were set to the conjugate 
of the likelihood density: Normal-Gamma prior for the mean and variance of the Gaussian components, 
symmetric Dirichlet prior over the HMM's state mixture's weights and symmetric Dirichlet prior over the 
acoustic units' weights.
For the case of the uninformative prior, the prior was set to be vague prior 
with one pseudo-observation \cite{Bishop2006:PRM} \footnote{Because of lack of space,
we have only given a rudimentary  description of the models. Note that the HMM model
was described at length in \cite{Ondel2016} whereas the full description of the SVAE 
model is yet to be published. However, the implementation of both models is available 
at https://github.com/amdtkdev/amdtk}. 

\section{Informative Prior}
\label{sec:informative_prior}

Bayesian Inference differs from other machine learning techniques by 
introducing a distribution $p(\GlobalParams)$ over the parameters of the
model. A major concern in Bayesian Inference is usually to define a prior that makes as little assumption 
as possible. Such a prior is usually known as uninformative prior. Having a completely uninformative prior has the practical 
advantage that the prior distribution will have a minimal impact on the outcome of the inference 
leading to a model which bases its prediction purely and solely on the data. 
In the present work, we aim at the opposite behavior, we wish our AUD 
model to learn phone-like units from the unlabeled speech data of a target
language given the knowledge that was previously accumulated from another
resourceful language. More formally, the original AUD model training consists in  estimate the \emph{a posteriori}
distribution of the parameters given the unlabeled speech data of a target language $\Matrix{X}_t$:

\begin{equation}
p(\GlobalParams | \Matrix{X}_t) = \frac{p(\Matrix{X}_t | \GlobalParams) p(\GlobalParams)}{p(\Matrix{X}_t)}
\end{equation} The parameters  are divided into two subgroups $\GlobalParams = \{\Params, 
\LatentVar_t\}$ where $\Params$ are the global parameters of the model, and $\LatentVar_t$ are the 
latent variables which, in our case, correspond to the sequences of  acoustic units. The global parameters
are separated into two independent subsets~: $\Params = \{ \Params_A, \Params_L\}$, corresponding to the 
acoustic parameters ($\Params_A$) and the "phonotactic" language model parameters ($\Params_L$). Replacing 
$\Matrix{\Params}$ and following the conditional independence of the variable induced by the model (see 
\cite{Ondel2016} for details) leads to:

\begin{equation}
\label{eq:bayes_noinf_prior}
p(\LatentVar_t, \Params | \Matrix{X}_t) \propto p(\Matrix{X}_t | \LatentVar_t, \Params_A) p(\LatentVar_t|\Params_L) p(\Params_L)p(\Params_A)
\end{equation}
If we further assume that we have at our disposal speech data in a different language 
than the target one, denoted $\Matrix{X_p}$, along with its phonetic transcription 
$\LatentVar_p$, it is then straightforward to show that:
\begin{align}
\label{eq:bayes_inf_prior}
p(\Params, \LatentVar_t | \Matrix{X}_t, \Matrix{X}_p, \LatentVar_p) &\propto  & p(\Matrix{X}_t | \LatentVar_t, \Params_A)p(\LatentVar_t | \Params_L)p(\Params_A|\Matrix{X}_p, \LatentVar_p)
\end{align}
which is the same as Eq. \ref{eq:bayes_noinf_prior} but for the distribution of the acoustic 
parameters which is based on the data of the resourceful language. In contrast of the term 
uninformative prior we denote $p(\Params_A | \Matrix{X}_p, \LatentVar_p)$ as an informative 
prior. As illustrated by Eq. \ref{eq:bayes_inf_prior}, a characteristic of Bayesian inference is that it 
naturally leads to a sequential inference. Therefore, model training can be summarized as:
\begin{itemize}
	\item given some prior data $\Matrix{X}_p$ from a resourceful language, estimate a posterior 
    distribution over the acoustic parameters $p(\Params_A|\Matrix{X}_p)$
    \item for a new unlabeled speech corpus, estimate the posterior distribution but considering
    the learned posterior distribution $p(\Params_A|\Matrix{X}_p)$ as a "prior".
\end{itemize}
Practically, the computation of the informative prior as well as the final posterior distribution is 
intractable and we seek for an approximation by means of the well known Variational Bayes Inference 
\cite{Jordan1999}. The approximate informative prior $q_1(\Params_A)$ is estimated by optimizing the 
variational lower bound of the evidence of the prior data $\Matrix{X}_p$:

\begin{equation}
  \begin{split}
    q_1^* = \argmax_{q_1} \; & \Exp_{q_1(\Params_A)} \big[ \ln 
        p(\Matrix{X}_p, \Params_A | \LatentParams_p,) \big] \\
        &  - \KL (q_1(\Params_A) || p(\Params_A))
  \end{split}
\end{equation}
where $\KL$ is the Kullback-Leibler divergence. Then, the posterior distribution of the parameters given
the target data $q_2(\Matrix{\LatentParams}_t, \Params_A, \Params_L)$ can be estimated by optimizing 
the evidence of the target data $\Matrix{X}_t$:

\begin{equation}
  \label{eq:vb_training}
  \begin{split}
    q_2^* = \argmax_{q_2} \; & \Exp_{q_2(\Matrix{\LatentParams}_t,
    	\Params_A, \Params_L)} \big[ \ln 
        p(\Matrix{X}_t, \Matrix{\LatentParams}_t, \Params_A, \Params_L) \big] \\
        &  - \KL (q_2(\Params_A) || q_1(\Params_A)) \\
        &  - \KL (q_2(\LatentParams_t, \Params_L) || p(\LatentParams_t, \Params_L))
  \end{split}
\end{equation}
Note that when the model is trained with an uninformative prior the loss function is 
the as in Eq. \ref{eq:vb_training} but with $p(\Matrix{\eta}_A)$ instead of the $q_1(\Matrix{\eta}_A)$. 
For the case of the uninformative prior, the Variational Bayes Inference was initialized as described in 
\cite{Ondel2016}. In the informative prior case, we initialized the algorithm by setting $q_2(\Params_A) = 
q_1(\Params_A)$.

\section{Experimental Setup}
\label{sec:setup}

\subsection{Corpora and acoustic features}
\label{subsec:database}
We used the Mboshi5K corpus \cite{mboshi-arxiv} as a test set for all the experiments reported here. {\embosi} (Bantu C25) is a typical Bantu language 
spoken in Congo-Brazzaville. It is one of the languages documented by the BULB (Breaking the Unwritten 
Language Barrier) project~\cite{addaBulbSLTU2016}. This speech dataset was collected 
following a real language documentation scenario, using  \ligaikuma \footnote{\url{http://lig-
aikuma.imag.fr}}, a mobile app specifically dedicated to fieldwork language documentation, which works both 
on android powered smartphones and tablets \cite{blachon2016}. The corpus is multilingual (5130 {\embosi} 
speech utterances aligned to French text) and contains linguists' transcriptions in {\embosi} (in the form 
of a non-standard graphemic form close to the language phonology). It is also enriched with 
automatic forced-alignment between speech and  transcriptions. The dataset is made available to the 
research community\footnote{It will be made available for free from ELRA, but its current version is online 
on: \url{https://github.com/besacier/mboshi-french-parallel-corpus}}. More details on this corpus can be 
found in \cite{mboshi-arxiv}. 

TIMIT is also used as an extra speech corpus to train the informative prior. We used two different set of 
features: the mean normalized MFCC + $\Delta$ + $\Delta\Delta$ generated by HTK and the Multilingual 
BottleNeck (MBN) features \cite{Grezl2014} trained on the Czech, German, Portuguese, Russian, Spanish, 
Turkish and Vietnamese data of the Global Phone database. 


\subsection{Acoustic unit discovery (AUD) evaluation}
\label{subsec:acoustic unit evaluation}

To evaluate our work we measured how the discovered units compared to the forced 
aligned phones in term of segmentation and information. The accuracy of the 
segmentation was measured in term of Precision, Recall and F-score. If a unit boundary occurs at the same time (+/- 10ms) of an actual 
phone boundary it is considered as a true positive, otherwise it is considered to be
a false positive. If no match is found with a true phone boundary, this is considered to be a false negative. The consistency of the units was evaluated
in term of normalized mutual information (NMI - see \cite{Ondel2016,Ondel2017,Ebbers2017} for details) which 
measures the statistical dependency between the units and the forced aligned phones. 
A NMI of 0 \%
means that the units are completely independent of the phones whereas a NMI of 
100 \% indicates that the actual phones could be retrieved without error given the
sequence of discovered units. 

\subsection{Extension to word discovery}
\label{subsec:word_discovery}

In order to provide an extrinsic metric to evaluate the quality of the acoustic units discovered by our different methods, we performed an unsupervised word segmentation task on the acoustic units sequences, and evaluated the accuracy of the discovered word boundaries. We also wanted to experiment using lattices as an input for the word segmentation task, instead of using single sequences of units, so as to better mitigate the uncertainty of the AUD task and provide a companion metric that would be more robust to noise.
A model capable of performing word segmentation both on lattices and text sequences was introduced by \cite{HeymannW2013}. Building on the work of \cite{Mochihashi09bayesian, Neubig12bayesian} they combine a nested hierarchical Pitman-Yor language model with a Weighted Finite State Transducer approach. Both for lattices and acoustic units sequences, we use the implementation of the authors with a bigram language model and a unigram character model\footnote{It would be more natural to use a 4-gram or an even higher order spelling model for word discovery, but we wanted to be able to validate our metric by matching it with the model of \cite{Goldwater09bayesian} ($dpseg$)  which implements a bigram language model based on a unigram model of characters (see details in Table \ref{tab:word-boundary}).}. 
	Word discovery is evaluated using the \textit{Boundary} metric from the \textit{Zero Resource Challenge 2017} \cite{ludusan2014} and \cite{zrc2017}. This metric measures the quality of a word segmentation and the discovered boundaries with respect to a gold corpus (Precision, Recall and F-score are computed).

\section{Results and Discussion}
\label{sec:results}

\begin{table}
  \centering
  \scalebox{.85}{
  \begin{tabular}{|c|c|c|c|c|}
      \hline 
      Features & Precision & Recall & F-score & NMI \\
      \hline
      MFCC &28.40 & $\mathbf{54.36}$ & 37.36 & 17.92 \\
      \hline
      MBN & 24.60 & 41.71 & 30.95 & 14.81 \\
      \hline
  \end{tabular}}
  \caption{AUD results of the baseline (HMM model with uninformative prior) - Mboshi5k corpus}
  \label{tab:vad_feat}
\end{table}

\begin{table*}[!t]
	\centering
    \scalebox{.7}{
  \begin{tabular}{|c|c|c||c|c|c||c|c|c|}
        \hline 
      Features & Model & Inf. Prior & Precision & Recall & F-score & Precision & Recall & F-score \\
      \cline{4-9} & & & \multicolumn{3}{|c||}{1-best Word Seg. }   &   \multicolumn{3}{|c|}{Lattices Word Seg. }  \\

      \hline 

      \hline 
      MFCC & HMM & no  & 28.8 &	74.5 &	41.5   &  29    & 75.9    & 41.9   \\
      MFCC & HMM & yes & 28.5 & $\mathbf{79.1}$ &	41.9   &  29.3 & $\mathbf{78.1}$ & 42.6   \\
      \hline
      MFCC & SVAE & no &  28.7 & 77.2 & 41.9    &  30   & 74.4 & 42.8\\
      MFCC & SVAE & yes & 29.3 & 73.1 & 41.8    &  30.4	& 69.6 & 42.3  \\
      \hline
      MBN & HMM & no   & $\mathbf{30.3}$  & 69   & 42.1    & 30.8  & 66.1 & 42 \\
      MBN & HMM & yes  & 29.2 & 67.5 & 40.8    & 29.9  & 67.8 & 41.5 \\
      \hline
      MBN & SVAE & no &  29.2 & 68.3 & 41      & 29.6  & 68.1 & 41.3 \\
      MBN & SVAE & yes & 29.8 & 73.4 & $\mathbf{42.4}$    & $\mathbf{30.9}$  & 72.2 & $\mathbf{43.3}$ \\
      \hline
  \end{tabular}}
  \caption{Precision, Recall and F-measure on word boundaries, using different AUD methods. Segmental DTW baseline \cite{aren} gave F-score of 19.3\% on the exact same corpus;
  $dpseg$ \cite{Goldwater09bayesian} was also used as a word segmentation baseline and gave similar (slightly lower) F-scores to \textit{1-best} (best config with $dpseg$ gave 42.5\%) - Mboshi5k corpus }
  \label{tab:word-boundary}
\end{table*}
First, we evaluated the standard HMM model with an uninformative prior (this will be our 
baseline) for the two different input features: MFCC (and derivatives) and MBN. Results are
shown in Table \ref{tab:vad_feat}. Surprisingly, the MBN features perform relatively poorly 
compared to the standard MFCC. These results are contradictory to those reported in \cite{Ondel2017}.
Two factors may explain this discrepancy: the Mboshi5k data being different from the training data
of the MBN neural network, the neural network may not generalize well. Another possibility may be that 
the initialization scheme of the model is not suitable for this type of features. 
Indeed, Variational 
Bayesian Inference algorithm converges only to a local optimum of the objective 
function and is therefore dependent of the initialization. We believe the 
second explanation is the more likely since, as we shall see shortly, the best 
results in term of word segmentation and NMI are eventually obtained with the MBN 
features when the inference is done with the informative prior. 
Next, we compared the HMM and the SVAE models when trained with an uninformative prior (lines with "Inf. 
Prior" set to "no" in Table \ref{tab:inf_prior}). The SVAE significantly improves the NMI and the 
precision showing that it extracts more consistent units than the HMM model. However, it also degrades the 
segmentation in terms of recall. 
%
%
We further investigated this behavior by looking at the duration of the units found by both models 
compared to the true phones (Table \ref{tab:unit_duration}). We observe that the SVAE model favors
longer units than the HMM model hence leading to fewer boundaries and consequently smaller recall. 
\begin{table}
  \centering
  \scalebox{.6}{
  \begin{tabular}{|c|c|c|}
      \hline 
      Features & Model & Average Unit duration (s)  \\
      \hline
      \multicolumn{2}{|c|}{phones} & 0.091 \\
      \hline 
      MFCC & HMM & 0.082 \\
      MFCC & SVAE &  0.096\\
      \hline
      MBN & HMM & 0.093 \\
      MBN & SVAE & 0.102 \\
      \hline
  \end{tabular}}
  \caption{Average duration of the (AUD) units (AUD) for the HMM and SVAE models trained with an uninformative prior. "phones"
  refers to the forced aligned phone reference.}
  \label{tab:unit_duration}
\end{table}

We then evaluated the effect of the informative prior on the acoustic unit discovery (Table 
\ref{tab:inf_prior}). On all 4 combinations (2 features sets $\times$ 2 models) we observe an 
improvement in terms of precision and NMI but a degradation of the recall. This result is 
encouraging since the informative prior was trained on English data (TIMIT) which is very different from Mboshi. Indeed, this suggests that even speech from an unrelated language can be of some help in the 
design of an ASR for a very low resource language. Finally, similarly to the SVAE/HMM case described above,
we found that the degradation of the recall is due to longer units discovered for models with 
an informative prior (numbers omitted due to lack of space).

\begin{table}
  \centering
  \scalebox{.6}{
  \begin{tabular}{|c|c|c||c|c|c|c|}
      \hline 
      Features & Model & Inf. Prior & Precision & Recall & F-score & NMI  \\
      \hline 
      MFCC & HMM & no & $28.4$ & $\mathbf{54.36}$ & $\mathbf{37.6}$ & 17.92  \\
      MFCC & HMM & yes & $29.88$ & $47.34$ & $36.64$ & $20.42$ \\
      \hline
      MFCC & SVAE & no & $30.1$ & $49.29$ &  $37.38$ & $21.03$ \\
      MFCC & SVAE & yes & $\mathbf{35.85}$ & 25.59 &  29.87 & 21.67 \\
      \hline
      MBN & HMM & no & $24.6$ &  $41.71$ & $30.95$ & 14.81 \\
      MBN & HMM & yes & $27.8$ & $36.58$ &  $31.56$ & $20.34$ \\
      \hline
      MBN & SVAE & no & $26.8$ & $41.51$ &  $32.57$ & $18.33$ \\
      MBN & SVAE & yes & $30.75$ & $37.94$ & $\mathbf{33.97}$ & $\mathbf{23.49}$ \\
      \hline
  \end{tabular}}
  \caption{Effect of the informative prior on AUD (phone boundary detection) - Mboshi5k corpus}
  \label{tab:inf_prior}
\end{table}
%
Word discovery results are given in Table \ref{tab:word-boundary} for the \textit{Boundary} metric 
 \cite{ludusan2014,zrc2017}. We observe  that i) the best word boundary detection (F-score) is obtained 
 with MBN features, an informative prior and the SVAE model;
  this confirms the results of table \ref{tab:inf_prior} and shows that better AUD leads to better word segmentation ii) word segmentation from AUD graph 
 \textit{Lattices} is slightly better than from flat  sequences of AUD symbols (\textit{1-best}); iii) our results outperform a pure speech based baseline based on segmental DTW \cite{aren} (F-score of 19.3\% on the exact same corpus).

\section{Conclusion}
\label{sec:conclusion}

We have conducted an analysis of the state-of-the-art Bayesian approach for acoustic unit discovery on a 
real case of low-resource language. This analysis was focused on the quality of the discovered units 
compared to the gold standard phone alignments. Outcomes of the 
analysis are i) the combination of neural network and Bayesian model (SVAE) yields a significant 
improvement in the AUD in term of consistency ii) Bayesian models can naturally embed information from a 
resourceful language and consequently improve the consistency of the discovered units.
Finally, we hope this work can serve as a baseline for future research on unsupervised 
acoustic unit discovery in very low resource scenarios. 

\section{Acknowledgements}
This work was started at JSALT 2017 in CMU, Pittsburgh, and was supported by JHU and CMU (via grants from Google, Microsoft, Amazon, Facebook, Apple), by the Czech Ministry of Education, Youth and Sports from the National Programme of Sustainability (NPU II) project "IT4Innovations excellence in science - LQ1602" and by the French ANR and the German DFG under grant ANR-14-CE35-0002 (BULB project). This work used the Extreme Science and Engineering Discovery Environment (NSF grant number OCI-1053575 and NSF award number ACI-1445606).

\bibliographystyle{IEEEbib}
\bibliography{refs}

\end{document}